\documentclass[lettersize,journal]{IEEEtran}
\usepackage{amsmath,amsfonts}
\usepackage{algorithmic}
\usepackage{algorithm}
\usepackage{array}
\usepackage[caption=false,font=normalsize,labelfont=sf,textfont=sf]{subfig}
\usepackage{textcomp}
\usepackage{stfloats}
\usepackage{url}
\usepackage{verbatim}
\usepackage{graphicx}
\usepackage{cite}
\usepackage{multirow}
\begin{document}

\title{WAVE: Wavelet-Controlled Coarse-to-Fine Guidance for Depth Super-Resolution}

\author{Tayyab~Nasir,%~\IEEEmembership{Student Member,~IEEE,}
        ~Daochang~Liu,%~\IEEEmembership{Member,~IEEE,}
        ~and~Ajmal~Saeed~Mian,%~\IEEEmembership{Senior Member,~IEEE}%
\thanks{%Manuscript received November 01, 2026.
\textit{(Corresponding author: Tayyab Nasir.)}}%
\thanks{The authors are with the Department of Computer Science and Software Engineering, The University of Western Australia,
Perth WA 6009, Australia (e-mail: tayyabnasir22@gmail.com;
tayyab.nasir@research.uwa.edu.au.}}

\maketitle

\begin{figure*}[!t]
\centering
\includegraphics[width=0.97\textwidth]{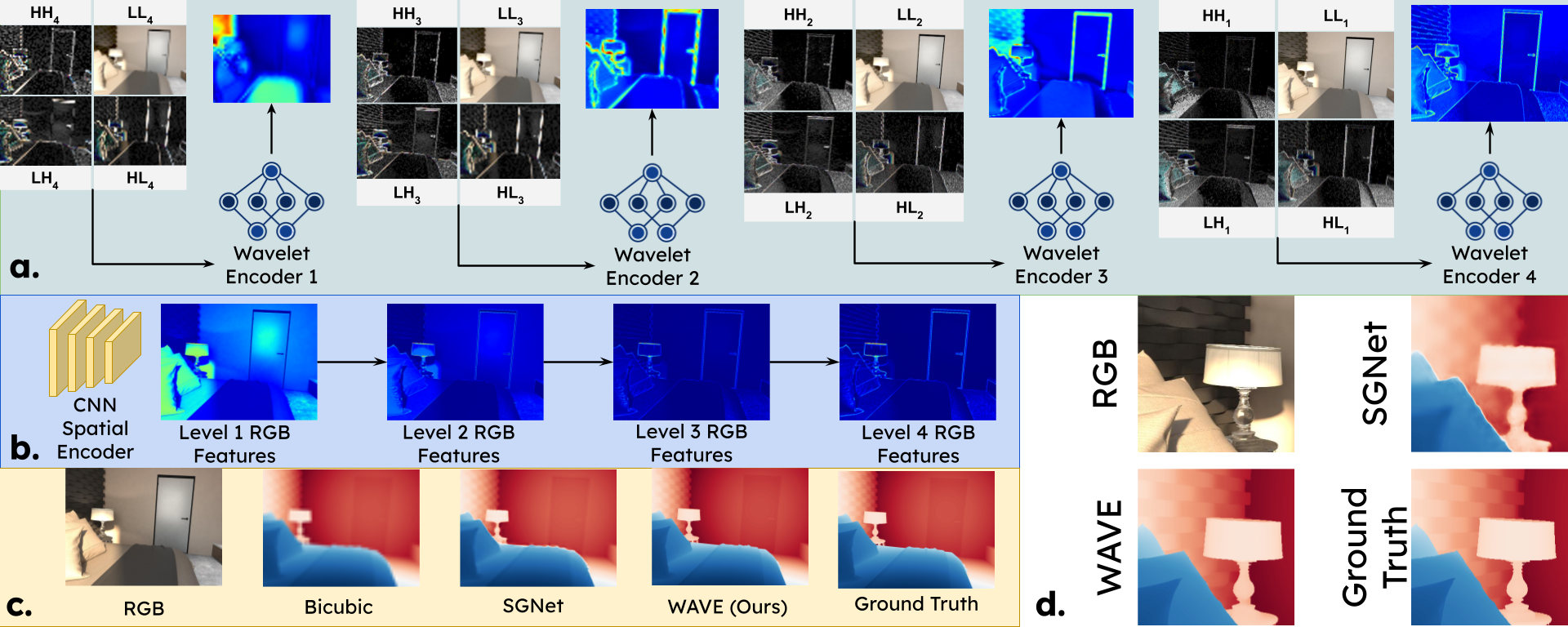} % Reduce the figure size so that it is slightly narrower than the column.
\caption{Coarse-to-fine spatial guidance in GDSR. \textbf{a.} WAVE uses ML-DWT and consumes the sub-bands in reverse of their generation order, i.e., coarsest first, restoring global structure before finer details. \textbf{b.} A conventional CNN-encoder for spatial feature extraction extracts RGB features directly, surfacing fine local cues in the earliest feature maps and recovering global structure only in deeper layers, where these early high-frequency cues must be suppressed later. \textbf{c.} Full-scene and \textbf{d.} zoomed comparisons show WAVE producing crisp and non-bleeding boundaries, with sharper contours and clean surfaces.}
\label{fig1}
\end{figure*}

\begin{abstract}
Guided depth super-resolution (GDSR) typically extracts RGB guidance features through convolutional hierarchies, inheriting their fine-to-coarse bias. Thus, low-level spatial cues surface in early layers, leaving the deeper layers to suppress those that do not correspond to true depth boundaries, which risks artifacts and blurred edges. The same fine-to-coarse bias persists in semantics-based methods that consume low-level tokens early and global tokens late. We present WAVE, which introduces a multi-level discrete wavelet transform (ML-DWT) as an explicit and interpretable feature-control mechanism, enabling a coarse-to-fine reconstruction by consuming sub-bands and semantic tokens in reverse of their generation order. WAVE further exploits these sub-bands to treat high- and low-frequency content separately, isolating the misleading RGB color and texture cues before any feature is learned so that their admission can be controlled explicitly, offering an intuitive alternative to the suppression learned implicitly by an opaque network. WAVE separates structure and detail reconstruction into dedicated modules that: i) model interactions within and across wavelet sub-bands, depth features, and semantic priors, ii) apply semantic gating to the high-frequency bands, and iii) fuse modalities through an invertible coupling mechanism that prevents collapse onto a single modality. Extensive experiments across multiple benchmarks demonstrate that WAVE matches or outperforms existing methods, with the largest gains at high upsampling factors, where low-resolution depth contains the least structure\footnote{Code available at: \url{https://github.com/tayyabnasir22/WAVE}}.
\end{abstract}


\section{Introduction}
Depth maps are fundamental to applications ranging from autonomous driving to augmented reality~\cite{yuan2023recurrent, yuan2023structure, zhong2023guided}. Although consumer RGB-D sensors provide an affordable way to capture the depth and RGB image of the same scene, the captured depth is at a lower spatial resolution compared to the RGB~\cite{yan2025ducos, wang2024scene}. Guided depth super-resolution (GDSR) addresses this limitation by exploiting the complementary information from the two modalities to reconstruct high-resolution depth maps. GDSR assumes that the RGB image and the depth map are two complementary geometric encodings of the same scene. However, it is widely observed in the literature that the RGB guide carries textures, color gradients, shadows, and illumination changes, providing structural cues that often do not coincide with true depth discontinuities. Such features, when used indiscriminately in the reconstruction process, generate artifacts, false depth discontinuities, and blurred object boundaries~\cite{kim2021deformable, nasir2026naima, he2021towards}. 

A large body of work has applied deep neural networks to GDSR, with recent approaches incorporating transformers, diffusion models, state-space models, and knowledge distillation, while convolutional neural networks (CNNs) remain the dominant backbone~\cite{de2022learning, hui2016depth, zuo2021mig, song2020channel, yang2022codon, zhong2021high, kim2021deformable, he2021towards, wu2026degmamba}. Recent studies have also introduced semantics, gradients, Fourier features, or multi-task supervision to mitigate artifacts induced by the misleading RGB guide cues~\cite{nasir2026naima, yan2025ducos, wang2024scene, metzger2023guided, wang2025dornet, wang2024sgnet, tang2021bridgenet}. In every case, however, the RGB guide enters the network through a learned encoder before any decision about which of its content is useful. Where frequency has been used at all, it has been either a single high-pass split of the guide~\cite{hui2016depth} or a transform applied to already-encoded features rather than to the guide image~\cite{zhao2022discrete, wang2024sgnet}. To our knowledge, none decomposes the guide into a multi-level, multi-band representation before encoding, which would allow misleading texture and color cues to be suppressed band-by-band and would let the network explicitly control which content, coarse structure or fine detail, is admitted at each stage of reconstruction.

A well-established property of convolutional feature extraction is that low-level cues such as edges and corners emerge in the early layers, while global structure forms progressively in deeper layers~\cite{yosinski2014transferable, zeiler2014visualizing}. Thus, CNN-based guide encoders process the RGB guidance in a fine-to-coarse manner, where the guide’s high-frequency texture and color cues are introduced early, deferring the removal of such noisy cues to the later layers where global structure finally emerges. Existing GDSR pipelines inherit this bias in three ways. 

First, many recent networks run a CNN backbone over the guide at a single working resolution and fuse its features into the depth stream, stage by stage~\cite{wang2024sgnet, wang2025dornet}. Since the RGB feature at each stage is the output of all the convolutional stages before it, the depth stream consumes the guide hierarchy in its native order, i.e., shallow, texture-dominated features first and deeper, structure-aware features last. Second, even a U-Net-style encoder-decoder is coarse-to-fine only in resolution, not in content. Some methods~\cite{metzger2023guided, kang2025c2pd} concatenate the RGB image with the upsampled depth and encode both with a U-Net, but the guide still passes through a convolutional encoder first, where texture enters at the very first layer and skip connections carry the shallowest features straight into the highest-resolution decoder stages. What reaches each decoder stage is therefore a learned feature whose frequency content is set by training rather than by design. Third, multi-scale guidance networks~\cite{hui2016depth, zhong2021high, zuo2021mig} upsample depth progressively and fuse CNN guide features at matching resolutions, so the deepest guide features are consumed first. This reverses the order of use, but it cannot guarantee the content. The guide encoder is trained jointly with reconstruction, so a coarse-resolution feature contains whatever lowered the training loss, including high-frequency cues, and what each stage receives is again decided implicitly by training rather than explicitly by design.

In the first two cases, the network must inject fine cues first and learn to suppress them later, a non-monotone process that is arguably harder to learn than a monotone one that fixes coarse structure first and refines detail last. The third case gets the order right but has no explicit handle on what is admitted at each stage.

Vision foundation models such as DINO~\cite{simeoni2025dinov3} and SAM~\cite{kirillov2023segment} have become a source of knowledge distillation for improving different vision tasks~\cite{liu2023one, zhang2025detect, yermakov2026deepfake, lin2025depth}. Their semantic representations are hierarchical as well, where shallow layers capture local appearance details while deeper layers encode increasingly global semantics. Recently, SPFNet~\cite{wang2024scene} and NAIMA~\cite{nasir2026naima} have distilled semantic knowledge from foundation models for improved GDSR. However, these methods largely retain the conventional fine-to-coarse information flow. For instance, NAIMA distills DINO tokens across layers but introduces the global deeper-layer tokens last, staying true to the fine-to-coarse ordering in semantic form as well.

A more natural ordering for depth reconstruction in GDSR, analogous to how a sculptor shapes the form of a figure before carving its detail, would reconstruct depth from coarse structure to fine detail. The discussion above shows that a learned guidance hierarchy, convolutional or semantic, is ill-suited to impose this ordering on its own. Enforcing a genuinely coarse-to-fine reconstruction therefore calls for a control mechanism that fixes both the order and the content of the guidance by construction rather than by training.

We propose WAVE, a multi-block, multi-branch architecture that challenges the prevailing fine-to-coarse paradigm. WAVE exploits Multi-Level Discrete Wavelet Transform~\cite{mallat1989theory} (ML-DWT) pyramid features together with DINO patch-level tokens, both inherently aligned in fine-to-coarse hierarchies, consuming them in reverse order to reconstruct depth from coarse structure to fine detail. Because the ML-DWT is a fixed transform rather than a learned encoder, the content admitted at each stage is set by construction, where the approximation band carries structure, and the detail bands carry edges and texture. At each stage, a divide-and-conquer strategy exploits the explicit decomposition of the wavelet transform, processing structural and high-frequency components through dedicated learnable branches within each reconstruction block. This offers a more intuitive refinement and filtering strategy than suppressing misleading RGB cues through an opaque end-to-end network. WAVE applies the ML-DWT to the RGB guide as an explicit mechanism for hierarchical, coarse-to-fine control and source-level noise suppression. WAVE also models interactions within and across frequency bands, between the bands and semantic tokens, and across modalities within the structural and detail branches. The cross-modal fusion step uses an invertible coupling~\cite{xiao2020invertible} rather than concatenate-and-project, so that neither modality can be discarded within the coupling. We verify empirically that the fused structural feature depends on the LL guidance, increasingly so at higher upsampling factors where more structure must come from RGB. Finally, contrary to prior semantic-guided approaches, which either rely on semantics alone (risking over-smoothed depth) or treat semantics as auxiliary structural cues only within the depth stream, WAVE employs semantics as a learnable gating mechanism that arbitrates which RGB content is admitted, in addition to guiding the depth branch. In summary, our main contributions are as follows:

\begin{itemize}
    \item We propose WAVE, which consumes the fine-to-coarse ML-DWT pyramid features and layer-wise DINO tokens in reverse, recovering global structure first and detail last with the content of each stage fixed by construction, countering the fine-to-coarse bias that prior methods inherit from CNN guidance extractors and from semantic-token pipelines that add global context only at later stages.
    \item To our knowledge, this is the first use of multi-level, multi-band wavelet decomposition of the RGB guide as an explicit guidance-control mechanism in GDSR, processing the structural, edge, and texture sub-bands independently, before any feature is learned, to address guide-induced noise at its source.
    \item We model three classes of interaction not previously unified in GDSR: among wavelet sub-bands, between wavelet features and semantics, and between the frequency-specific guidance and depth stream. Within these, learnable semantic gating modulates the high-frequency edge and texture bands, suppressing texture-copying artifacts while preserving geometrically meaningful details.
    \item We repurpose the IRN-style bijective coupling for cross-modal fusion, whose information-preserving form keeps both modalities recoverable within the coupling and discourages collapse onto a single modality, unlike prior work that uses invertible blocks only as spatial-frequency bridges within a single stream. 
    \item Beyond the direct injection of semantics into the depth stream, we adapt the frozen DINO via a low-rank residual on its output tokens rather than its weights (following DiReFT/LoRA~\cite{wu2024reft, hu2022lora}), obtaining task-specific semantics for GDSR without backbone fine-tuning or an explicit semantic loss. 
\end{itemize}

The role of each component is detailed in the Methodology. Extensive experiments across multiple benchmarks and scale factors show that WAVE performs particularly well at extreme upsampling, where the low-resolution input retains the least structure and coarse-to-fine reconstruction contributes most. At 32$\times$, WAVE attains a reduced RMSE over the next-best method C2PD (RGB-D-D: 3.77 vs.\ 4.11; NYU\_v2: 7.90 vs.\ 9.03; Average: 6.01 vs.\ 6.33).

\section{Literature Review}
Acquiring high-quality depth maps remains expensive and hardware-constrained compared to their RGB counterparts~\cite{ariav2022depth}, which has motivated a large body of work on guided depth super-resolution, where a high-resolution RGB image of the same scene guides the recovery of a high-quality depth map from its low-resolution version~\cite{zhong2023guided}. Deep neural networks have become the standard approach for GDSR, where early works framed the task as deep joint image filtering~\cite{li2016deep, li2019joint, kim2021deformable} or employed convolutional architectures to extract and fuse features from the low-resolution depth and the high-resolution RGB guide~\cite{zuo2021mig, hui2016depth, deng2020deep, de2022learning, tang2021joint}, later extending to real-world degradations~\cite{he2021towards} and more diverse formulations such as anisotropic diffusion~\cite{metzger2023guided}, continuity-constrained deformation~\cite{kang2025c2pd}, and state-space models~\cite{wu2026degmamba}. The guided formulation, however, carries an inherent weakness rooted in the RGB guide itself, i.e., textures, color gradients, shadows, and illumination patterns frequently do not coincide with true geometric discontinuities, and when transferred indiscriminately, they produce false depth edges, texture-copying artifacts, and blurred object boundaries~\cite{he2021towards, zhao2022discrete, nasir2026naima}. 

A substantial line of work mitigates this guidance-induced noise through selective feature integration, including attention-based fusion that models cross-modal correlation and suppresses depth-irrelevant responses~\cite{zhong2021high, yang2022codon, song2020channel, zhong2023deep, shi2022symmetric}, degradation-aware or structure-regularized fusion~\cite{wang2025dornet, yuan2023recurrent, yuan2023structure}, and auxiliary or multi-task supervision such as joint learning with monocular depth estimation~\cite{tang2021bridgenet} or depth completion~\cite{yan2022learning}. Semantic priors from foundation models have recently been distilled into GDSR~\cite{wang2024scene, yan2025ducos, nasir2026naima}. A smaller set of methods reasons about the guide in the frequency domain. MSG-Net~\cite{hui2016depth} applies a single high-pass filter to the guide, FDSR~\cite{he2021towards} learns an adaptive high and low frequency split, SGNet~\cite{wang2024sgnet} uses the gradient of the guide and a Fourier transform of its features, DCTNet~\cite{zhao2022discrete} applies a discrete cosine transform to already-encoded features, and IPPNet~\cite{chen2024intrinsic} learns depth-conditioned phase filters on encoded RGB features. In each case, the frequency split is either a single band or a learned operation on features, and the RGB guide still enters the network through a learned encoder.

Across all of these strategies, one structural issue persists. Whether convolutional or token-based, the backbones that consume the RGB guide share the same layer-wise behavior, i.e., high-frequency local cues such as edges and textures surface in the earliest layers, while global scene structure emerges only in the deeper ones~\cite{yosinski2014transferable, zeiler2014visualizing}. Networks that fuse these features stage by stage inject the misleading high-frequency content first and must learn to suppress it later, where global context is available. Even the multi-scale and encoder-decoder designs that fuse guidance in a coarse-to-fine schedule~\cite{hui2016depth, zhong2021high, zuo2021mig} are not truly coarse-to-fine, because the RGB image still passes through a convolutional encoder and what reaches each fusion stage is an implicit feature of that encoder. The coarse-level feature is not a coarse version of the guide but whatever the encoder learned to compute from the full-resolution image, texture included, and the fine-level feature is whatever a shallow layer produced before any global context existed. Reordering the fusion of these features changes when they are used, not what they contain, and nothing in the architecture guarantees that the coarse stage carries only low-frequency content or that the fine stage receives only detail. To our knowledge, no prior learning-based GDSR method partitions the raw guide in advance, assigning its frequency bands to reconstruction scales and consuming them in a set coarse-to-fine order.

WAVE places a fixed, parameter-free wavelet transform~\cite{mallat1989theory} in front of the guidance encoders. The RGB guide is decomposed before any learning takes place, so the structure band consumed at the coarse stage and the detail bands consumed at the fine stages are defined by the transform rather than by the network, and the decomposition is the same for every input and every training run. Each learned encoder then operates on a single band at a single scale, and the content admitted at each reconstruction stage can be read off directly from the decomposition. Coarse-to-fine is thus a property of the signal admitted at each stage rather than an expected property of implicit features.

\section{Methodology}
\begin{figure*}[t]
\centering
\includegraphics[width=0.95\textwidth]{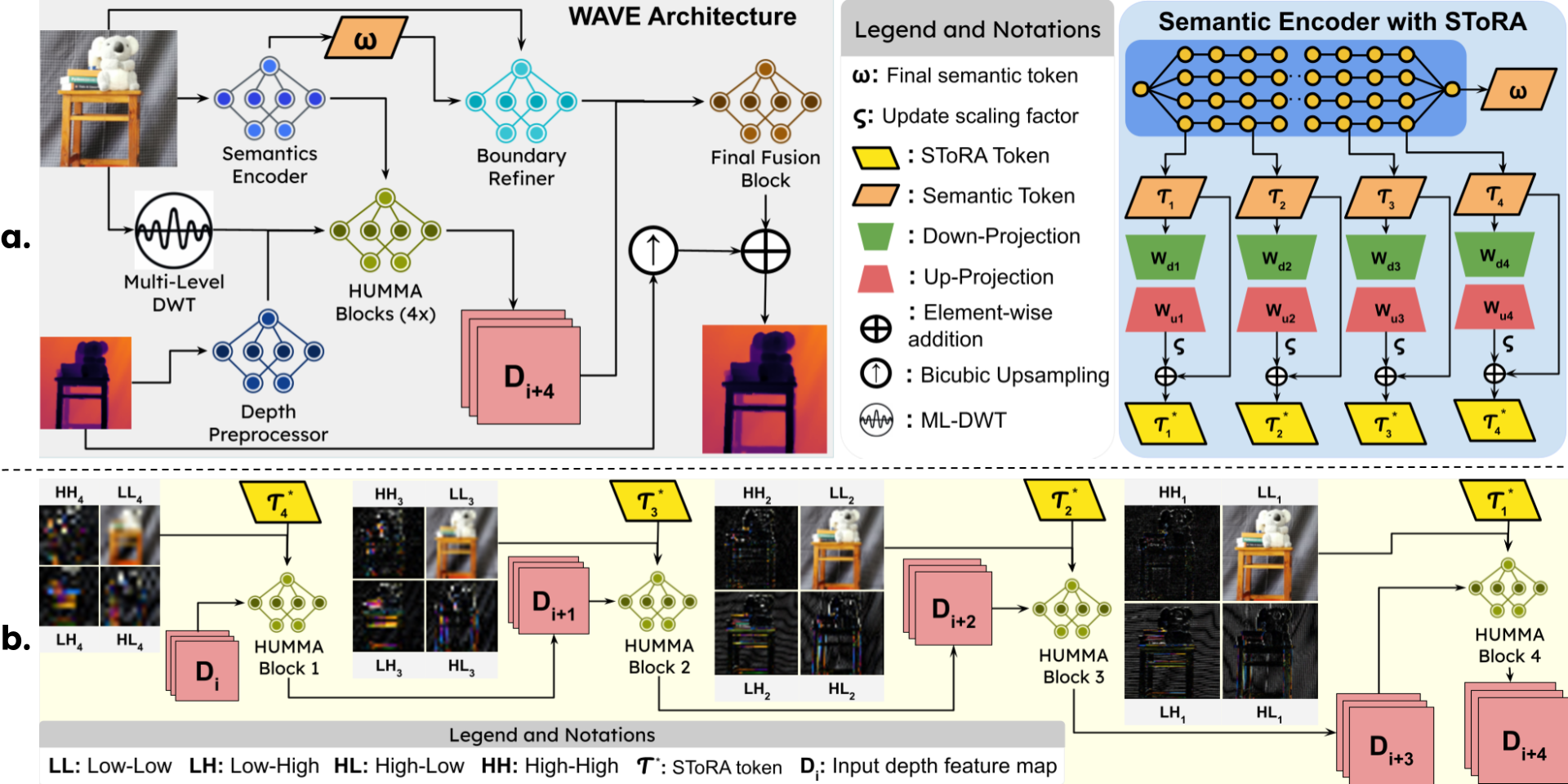}
\caption{Overview of the WAVE architecture and the coarse-to-fine HUMMA reconstruction blocks pipeline. \textbf{a.} ML-DWT decomposes the RGB guide into a fine-to-coarse sub-band pyramid, DINOv3 produces semantic tokens adapted by SToRA, and the Depth Preprocessor encodes the low-resolution depth. \textbf{b.} Four HUMMA blocks progressively reconstruct and upsample the depth feature maps using the coarse-to-fine sub-bands and semantic tokens, and the Final Fusion Block applies a global correction guided by semantic boundaries.}
\label{fig2}
\end{figure*}

Figure~\ref{fig2} presents an overview of our proposed WAVE architecture’s coarse-to-fine super-resolution pipeline. The following subsections present the role and implementation of different modules that make this explicit coarse-to-fine reconstruction possible for GDSR.

\subsection{WAVE - Coarse-to-Fine Reconstruction}
WAVE processes two input modalities: the low-resolution depth $D_{lr} \in \mathbb{R}^{1 \times H/s \times W/s}$ and the RGB image $I \in \mathbb{R}^{3 \times H \times W}$, where $s$ is the scaling factor. Unlike conventional methods that pass RGB directly through the network layers, we first decompose $I$ into approximation and detail sub-bands by applying a four-level Haar wavelet decomposition via Mallat's recursive multiresolution algorithm~\cite{mallat1989theory}. The transform is applied recursively to the low-frequency band, producing a multi-resolution pyramid in which each iteration halves the spatial resolution and the retained structure grows progressively coarser. Each level $l \in \{1,2,3,4\}$ thus yields one approximation band $LL_{l}$ which retains a smooth structural approximation of the scene, and three detail bands $LH_{l}, HL_{l}, HH_{l}$ which carry fine details such as edges and texture. This not only decomposes the guidance signal by frequency content, separating smooth structure from the high-frequency bands that carry edges and textures, but also provides this separation at multiple levels, with local detail at shallow levels and global structure at deep ones. The result is explicit control over which RGB content is admitted at each scale, fixed by the transform before any feature is learned. This serves as the mechanism for both controlling misleading RGB cues at the input and enforcing a coarse-to-fine reconstruction, neither of which arises naturally from the implicit fine-to-coarse hierarchy of CNN-based extractors~\cite{yosinski2014transferable}.

\textbf{SToRA:} We use DINOv3~\cite{simeoni2025dinov3} (ViT-S/16) to extract patch tokens $\{\tau_{l}\}_{l=1}^{4}$ from 4 layers (2, 5, 9, 11), each capturing a different level of semantic abstraction, supplying semantic priors from $I$ as a derived auxiliary guiding modality. DINOv3 is kept frozen, and we use a bias-free, DiReFT~\cite{wu2024reft}-inspired module for lightweight, task-specific adaptation of the tokens, rather than fine-tuning the backbone or resorting to an auxiliary semantic loss. For the token $\tau_{l} \in \mathbb{R}^{C}$ at level $l$, the low-rank residual is applied as:

{\small
\begin{equation}
    \tau_{l}^{*} = \tau_{l} + \varsigma  W_{u}(W_{d}\tau_{l}),
\end{equation}
}

\noindent where, $W_{u} \in \mathbb{R}^{C \times r}$ and $W_{d} \in \mathbb{R}^{r \times C}$ are low-rank projection matrices, $r << C$, and $\varsigma=\alpha/r$ scales the update. We zero-initialize $W_{u}$ so that adaptation begins with the pretrained representation and gradually deviates during training. This efficiently modulates token features for task-specific learning, using only $2Cr$ parameters per adapter compared with a full $C \times C$ transform. We call this lightweight calibration Semantic Token Residual Adapter (SToRA), which helps avoid overwriting the rich semantics encoded in the token while nudging it with task-specific information. An ablation of the model without this module is provided in the Ablations section.

\subsection{HUMMA Block}
\begin{figure}[!htb]
\centering
\includegraphics[width=0.97\columnwidth]{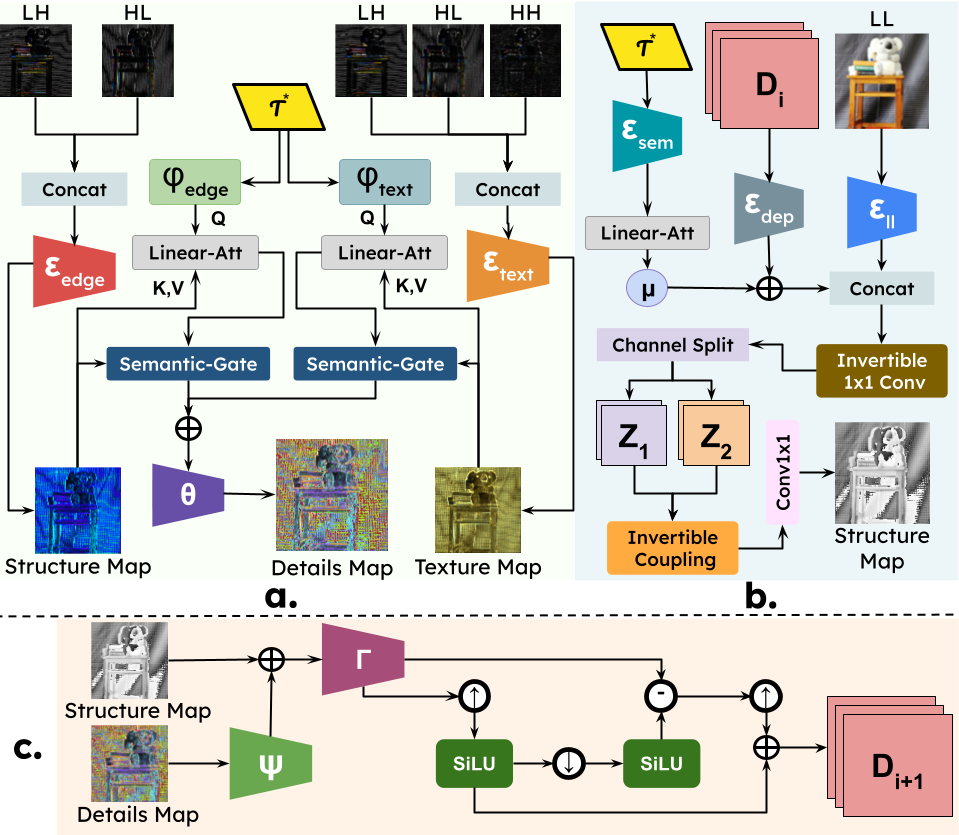} 
\caption{Overview of the HUMMA block. \textbf{a.} Detail branch processes the high-frequency sub-bands along edge and texture paths, each modulated by the SToRA token via semantic gating. The two feature maps are merged to produce a detail map. \textbf{b.} Structure branch relates the low-frequency depth features, LL approximation, and SToRA tokens through learnable projections. The IRN-style coupling bijectively mixes the modalities, discouraging collapse onto either. \textbf{c.} The Reconstruction block aligns and residually adds the details to the structure map, followed by back-projection for $2 \times$ upsampling to produce the next refined depth feature map.}
\label{fig3}
\end{figure}

HUMMA (Hierarchical Upsampling with Multi-band Multi-resolution Approximation), presented in Figure~\ref {fig3}, is the fundamental building block that produces a refined depth feature map from: i) the previous, lower-resolution depth features, ii) the ML-DWT sub-bands at the matching pyramid level, and iii) the semantic token with abstraction corresponding to the current level. The refined output depth feature map is upscaled by a factor of $2 \times$. To invert the implicit fine-to-coarse behavior of CNN extractors, the blocks operate on the DWT and DINOv3 hierarchies in reverse, towards progressively higher-resolution, finer-detail representations. Each HUMMA block splits the computation into a structure branch and a detail branch, so that misleading RGB cues are explicitly controlled for each frequency regime.

\textbf{Details Branch:} Recovers edge and texture details while suppressing high-frequency responses that do not correspond to true geometric boundaries. Rather than learning what detail is from an opaque feature map, the DWT provides an explicit handle over the three high-frequency sub-bands $LH, HL, HH \in \mathbb{R}^{3 \times h \times w}$ (each), where $h \times w = H / 2^{l} \times W / 2^{l}$, allowing WAVE to process edge and texture cues along two separate paths before merging them. Axis-aligned edge energy is concentrated in the horizontal and vertical bands, so we first concatenate and project the $LH$ and $HL$, collapsing them to a single $c$-channel feature map, and pass them through the edge encoder $\varepsilon_{edge}$, which is a short stack of $1 \times 1$ and $3 \times 3$ convolutions in which $SiLU(x) = x \cdot \sigma(x)$ acts as a self-gating nonlinearity~\cite{elfwing2018sigmoid}:

{\small
\begin{equation}
    F_{edge} = \varepsilon_{edge}([LH \mid HL]) \in \mathbb{R}^{c \times h \times w},
\end{equation}
}

\noindent where $c$ is our projection channel dimension for level $l^{*}$ such that $l^{*} = 1$ operates on the coarsest wavelet (and token) representation $l=4$, while $l^{*} = 4$ corresponds to the finest representation $l=1$. Texture additionally requires the diagonal band, so the texture branch collapses all three bands and uses $\varepsilon_{text}$ that interleaves dilated convolutions with a standard $3 \times 3$ fill convolution, enlarging the receptive field to capture multi-scale texture while $SiLU$ again self-gates:

{\small
\begin{equation}
    F_{text} = \varepsilon_{text}([LH \mid HL \mid HH]) \in \mathbb{R}^{c \times h \times w}.
\end{equation}
}

Hence, this way we learn intra-band relations, exploiting them to encode edge and texture features matched to their band statistics, instead of being entangled in a single guidance feature. The learnable interactions among bands are then followed by learnable interactions with the semantics to align the high-frequency features with true semantic boundaries. The SToRA token $\tau^{*}$ for the corresponding level is passed through different projections ($\varphi_{edge}$ and $\varphi_{text}$) to align with the feature space. 

{\small
\begin{equation}
    E_{guide} = \varphi_{edge}(\tau^{*}), \quad T_{guide} = \varphi_{text}(\tau^{*}).
\end{equation}
}

Each projected semantic guide then queries a global summary of its corresponding feature map through linear cross-attention~\cite{katharopoulos2020transformers}, selecting relevant details, operating at linear $\mathcal{O}(N)$ cost (N is the number of spatial positions $h \times w$):
{\small
\begin{equation}
\begin{gathered}
    Attn(Q, K, V) = \phi(Q)(\phi(K)^{\top}V),  \\
    \phi(x) = ELU(x) + 1. \\
    \tilde{F}_{edge} = Attn(E_{guide}, F_{edge}, F_{edge}),\\
    \tilde{F}_{text} = Attn(T_{guide}, F_{text}, F_{text}).
\end{gathered}
\end{equation}
}

\textbf{Learned Gated Semantic Mixing:} The semantically-modulated features are combined with the raw features in each path through learned coefficients $\gamma_{edge}$ and $\gamma_{text}$, initialized to zero so that each path relies entirely on its raw edge/texture features at the start of training, with the semantic contribution introduced gradually as training progresses:

{\small
\begin{equation}
\begin{gathered}
    F_{edge}^{*} = (1-\gamma_{edge})F_{edge} + \gamma_{edge}\tilde{F}_{edge} \\
    F_{text}^{*} = (1-\gamma_{text})F_{text} + \gamma_{text}\tilde{F}_{text}.
\end{gathered}
\end{equation}
}

This deferred reliance on semantics lets the network first learn stable band representations before the semantic guide reweights high-frequency responses toward those consistent with true geometric boundaries. The outputs of the two paths are summed and passed through a final refinement projection $\theta$ to produce the high-frequency detail feature map, subsequently used to sharpen the structural depth feature and to prevent over-smoothed depth:

{\small
\begin{equation}
    F_{details} = \theta(F_{edge}^{*} + F_{text}^{*}).
\end{equation}
}

\textbf{Structure Branch:} The relationship among three modalities, i.e., the low-frequency depth representation $D$, the global structural cues in the LL band, and the SToRA token $\tau^{*}$, is learned through a series of projections. The explicit control over the inputs, which is limited to a smooth, low-frequency structure of the scene while withholding the high-frequency appearance detail, helps this branch learn geometry without being distracted by dominating texture. Each modality is first brought to a common latent space through learned projections, with self-gating achieved through $SiLU$, adaptively suppressing irrelevant responses per stream. Let $E_{d}, E_{l}, S_{guide} \in \mathbb{R}^{c \times h \times w}$ denote the aligned representations corresponding to the encoded depth feature map, LL-subband, and semantic features, respectively. The relation between depth modality $D$ and semantics $S_{guide}$ is established for refining the depth map with semantic cues. To actively pull the semantic content relevant to the depth structure, linear cross-attention is applied using the depth feature map as a query:

{\small
\begin{equation}
\begin{gathered}
    S_{dep} = \phi(Q)(\phi(K)^{\top}V), \\
    Q = D, 
    K = V = S_{guide}.
\end{gathered}
\end{equation}
}

The query/key assignment is intentionally reversed from the detail branch: there, the semantic token queries the feature maps to form a global summary that gates local bands, while here, the depth map queries the semantics to pull structure-relevant content. The related semantics refine the depth feature map as a learnable-gated residual:

{\small
\begin{equation}
    E_{d}^{*} = E_{d} + \mu S_{dep} \in \mathbb{R}^{c \times h \times w},
\end{equation}
}

\noindent where $\mu$ is initialized to zero, so the branch begins from the raw depth feature and injects semantic correction only as training progresses, the same deferred-reliance gating mechanism used before. 

\textbf{Cross-Modal Coupling:} We adopt an IRN-style~\cite{xiao2020invertible} invertible coupling to achieve a mutually conditioned mix of structural cues from the semantically rich depth feature and the encoded photometric information from the LL approximation band. The semantically-refined depth feature $E_{d}^{*}$ and aligned LL representation $E_{l}$ are concatenated and mixed across channels by an LU-parameterized invertible $1 \times 1$ convolution, producing $Z \in \mathbb{R}^{2c \times h \times w}$. This mixed representation is then split into two halves $Z_{1}, Z_{2} \in \mathbb{R}^{c \times h \times w}$, which are coupled by an affine transform in which each half conditions the other:

{\small
\begin{equation}
\begin{gathered}
    Y_{1} = Z_{1} + F(Z_{2}), \quad
    S = 2\sigma(H(Y_{1})) - 1, \\
    Y_{2} = Z_{2} \odot \exp(S) + G(Y_{1}), \\
\end{gathered}
\end{equation}
}

\noindent where $F$, $G$, and $H$ are learned convolutional subnetworks, $\sigma$ is the sigmoid, and $\odot$ denotes the Hadamard product. Because the coupling is invertible by construction, it discourages the fusion from discarding either modality, in contrast to a plain concatenation-and-projection that can freely suppress one stream, which we verify empirically in Section~\ref{sec:modality} by zeroing the LL input at inference. The fused representation is passed through a separate learned $1 \times 1$ convolution yielding the structural feature $E_{struct} \in \mathbb{R}^{c \times h \times w}$ that matches the dimension expected by the subsequent detail injection and upsampling.

The structural depth feature, reconstructed from the low-frequency content, is then injected with the high-frequency details recovered by the detail branch. The details are aligned to the structural feature through a learned projection $\psi$, and then added residually, scaled by a learnable coefficient $\rho$ initialized to zero, and the result is harmonized by a residual group of channel-attention blocks $\Gamma$:

{\small
\begin{equation}
\begin{gathered}
    \tilde{F}_{details} = \psi(F_{details}), \\ 
    \tilde{D} = \Gamma(E_{struct} + \rho \tilde{F}_{details}).
\end{gathered}
\end{equation}
}

The additive form and the zero-initialized residual keep the reconstructed structure intact while the details sharpen it only where warranted, progressively through the training in a controlled manner. Finally, a DBP-styled~\cite{haris2018deep} back-projection block upsamples the depth feature map to $2 \times$:

{\small
\begin{equation}
\begin{gathered}
    H_{0} = \mathrm{SiLU}(\uparrow \tilde{D}), \quad
    L_{0} = \mathrm{SiLU}(\downarrow H_{0}), \\
    E_{err} = L_{0} - \tilde{D}, \quad
    H_{1} = \uparrow E_{err}, \quad
    \tilde{D}^{\uparrow} = H_{0} + H_{1},
\end{gathered}
\end{equation}
}

\noindent where $\uparrow$ and $\downarrow$ denote the learned up- and down-projections, implemented as transposed and strided convolutions at $2 \times$ scale.

\subsection{Boundary Refinement and Optimization}
To add global, object-level contour sharpening, the coarse-to-fine reconstructed feature map is subjected to a final refinement, guided by a semantic boundary derived from the source RGB. Self-supervised vision transformer features carry implicit scene-layout and object-boundary information accessible in the final-layer tokens~\cite{simeoni2025dinov3, caron2021emerging, oquab2023dinov2}, which we exploit by computing the cosine similarity between each patch and its horizontal and vertical neighbors~\cite{simeoni2025dinov3}, reading out boundary strength as the local dissimilarity. For the $l_{2}$-normalized final patch token $\omega$:

{\small
\begin{equation}
\begin{gathered}
    c_x = \langle \omega_{i,j}, \omega_{i,j-1}\rangle, \quad
    c_y = \langle \omega_{i,j}, \omega_{i-1,j}\rangle, \\
    B = \sqrt{(1-c_x)^2 + (1-c_y)^2}.
\end{gathered}
\end{equation}
}

Since these dense similarities are known to be spatially noisy~\cite{simeoni2025dinov3}, $B$ is smoothed with a Gaussian kernel and mean-normalized for scale invariance, yielding the soft semantic boundary $\tilde{B}$. The boundary $\tilde{B}$, the projected input RGB, and the depth map from the last HUMMA block, each at the same spatial resolution $H \times W$, are passed through a learnable fusion network to produce the final high-resolution depth $D_{hr} \in \mathbb{R}^{1 \times H \times W}$. For optimization we use modified $L_{1}$ loss~\cite{nasir2026implicit}.

\section{Experiments}
\subsection{Datasets and Evaluation}
We adopt two established GDSR benchmarking protocols~\cite{kang2025c2pd, yan2025ducos, zhong2026dual} that differ in training data and evaluation sets. The first trains on HYPERSIM~\cite{roberts2021hypersim} and tests on Middlebury~\cite{scharstein2003high}, Lu~\cite{lu2014depth}, NYU\_v2~\cite{silberman2012indoor}, RGB-D-D~\cite{he2021towards}, and TOFDSR~\cite{yan2024tri}. The second trains on NYU\_v2 and tests on NYU\_v2, RGB-D-D, Middlebury, Lu, and DIML~\cite{cho2021diml}. Under each protocol, we follow the exact training, augmentation, and evaluation settings of the works from which baseline numbers are taken, i.e., DuCos~\cite{yan2025ducos} for HYPERSIM and SPFNet~\cite{wang2024scene} and C2PD~\cite{kang2025c2pd} for NYU\_v2, enabling a direct comparison with existing methods~\cite{li2016deep, li2019joint, deng2020deep, kim2021deformable, tang2021joint, zhao2022discrete, metzger2023guided, zhao2023spherical, wang2024sgnet, wang2025dornet, nasir2026naima}. We report RMSE in centimeters (lower is better) at $8\times$ and $16\times$, and additionally at $32\times$ to assess extreme upsampling, following~\cite{wang2024scene, kang2025c2pd}.

\textbf{Implementation details:} All WAVE models are trained for 200 epochs using Adam with an initial learning rate of $1\mathrm{e}{-4}$. One model is trained per protocol and scaling factor, with the DINOv3 ViT-S/16 backbone frozen throughout. All experiments run on a single NVIDIA RTX 4090. Code and trained weights are publicly available. Table~\ref{tab:params} compares parameters, FLOPs, and peak memory across methods.

\subsection{Benchmarking}
Tables~\ref{tab:largerscale},~\ref{tab:hypersim}, and~\ref{tab:nyu_v2} report the results, where WAVE shows competitive performance across all settings. Its margins are largest at higher upsampling factors, consistent with our hypothesis that coarse-to-fine reconstruction helps most when the low-resolution input retains the least structure. Under the HYPERSIM protocol (Table~\ref{tab:hypersim}), WAVE attains the best average RMSE at both scales, with the largest reductions on the out-of-domain TOFDSR set (5.33 vs.\ 6.85 at $16\times$). Under the NYU\_v2 protocol at $8\times$ and $16\times$ (Table~\ref{tab:nyu_v2})\footnote{LapNet~\cite{zhong2026dual} and IPPNet~\cite{chen2024intrinsic} are not compared as neither had released code or weights at the time of writing. Re-implementing from the paper alone risks misrepresenting the method, and without weights no results can be obtained for the qualitative, complexity, and misalignment analyses.}, the in-domain test set is saturated (leading methods within $\sim$ 0.1--0.2cm RMSE), and WAVE is competitive rather than dominant there. WAVE achieves the lowest RMSE at both scales on RGB-D-D (405 pairs) and DIML (503 pairs), the two out-of-domain and larger datasets, and hence the more reliable indicators of generalization than Middlebury (30 pairs) and Lu (6 pairs). At $32\times$ (Table~\ref{tab:largerscale}), WAVE obtains the best average and the lowest RMSE on three of the four test sets, trailing C2PD only on Lu.

\begin{table}[!htb]
    \caption {$32 \times$ results under the NYU\_v2 training protocol, evaluated on RGB-D-D, NYU\_v2, Middlebury, and Lu. Baseline numbers are taken from~\cite{kang2025c2pd}. Best in \textbf{bold}, second-best \underline{underlined}.}
	\centering
    \footnotesize
    \setlength{\tabcolsep}{5pt}
	\begin{tabular}{l|c|c|c|c|c}
    \hline
		Method
        & \textbf{RGB-D-D}
        & \textbf{NYU\_v2}
        & \textbf{M-bury}
        & \textbf{Lu}
        & \textbf{Average} \\
		\hline

        DJFR   & 6.48 & 14.12 & 8.57 & 9.94 & 9.78 \\
        CUNet  & 7.75 & 15.95 & 9.23 & 10.97 & 10.98 \\
        DKN    & 5.97 & 12.46 & 7.76 & 8.98 & 8.79 \\
        FDSR   & 5.08 & 14.19 & 7.26 & 8.62 & 8.79 \\
        DCTNet & 5.99 & 13.33 & 8.22 & 9.28 & 9.21 \\
        SGNet  & 5.15 & 11.24 & 6.79 & 7.23 & 7.60 \\
        DORNet & 5.05 & 10.97 & 6.87 & 8.28 & 7.79 \\
        C2PD & \underline{4.11} & \underline{9.03} & \underline{5.68} & \textbf{6.48} & \underline{6.33} \\
        \hline
        \textbf{WAVE}   & \textbf{3.77} & \textbf{7.90} & \textbf{5.42} & \underline{6.96} & \textbf{6.01} \\

        \hline
	\end{tabular}
	
	\label{tab:largerscale}
\end{table}

\begin{table*}[!htb]
    \caption {Hypersim training protocol at $8 \times$ and $16 \times$, evaluated on RGB-D-D, TOFDSR, NYU\_v2, Middlebury, and Lu. Baselines taken from~\cite{yan2025ducos}. Best in \textbf{bold}, second-best \underline{underlined}}
	\centering
    \footnotesize
    \setlength{\tabcolsep}{9pt}
	\begin{tabular}{l|cc|cc|cc|cc|cc|cc}
    \hline
		\multirow{2}{*}{Method} 
        & \multicolumn{2}{c|}{\textbf{RGB-D-D}}
        & \multicolumn{2}{c|}{\textbf{TOFDSR}}
        & \multicolumn{2}{c|}{\textbf{NYU\_v2}}
        & \multicolumn{2}{c|}{\textbf{M-bury}}
        & \multicolumn{2}{c|}{\textbf{Lu}}
        & \multicolumn{2}{c}{\textbf{Average}} \\
        
        & $8 \times$ & $16 \times$ 
        & $8 \times$ & $16 \times$
        & $8 \times$ & $16 \times$  
        & $8 \times$ & $16 \times$
        & $8 \times$ & $16 \times$
        & $8 \times$ & $16 \times$
        \\
		\hline
                
    DJF    & 2.58 & 4.46 & 5.59 & 8.19 & 5.56 & 9.82 & 3.09 & 5.50 & 3.58 & 6.53 & 4.08 & 6.90 \\
    DJFR   & 2.61 & 4.36 & 5.11 & 8.06 & 5.20 & 9.50 & 2.82 & 5.16 & 3.24 & 6.46 & 3.80 & 6.71 \\
    CUNet  & 2.35 & 3.81 & 5.14 & 7.36 & 5.50 & 8.63 & 2.86 & 4.72 & 2.85 & 5.63 & 3.74 & 6.03 \\
    FDKN   & 2.25 & 3.71 & 4.40 & 7.16 & 4.93 & 7.97 & 2.51 & 4.42 & \underline{2.67} & 5.48 & 3.35 & 5.75 \\
    DKN    & 2.33 & 3.70 & 4.54 & 7.24 & 4.88 & 7.70 & 2.43 & 4.17 & 2.88 & 5.44 & 3.41 & 5.65 \\
    FDSR   & 2.25 & \underline{3.44} & \underline{4.28} & \underline{6.85} & \underline{4.82} & \underline{7.29} & 2.41 & 3.97 & 2.69 & 5.23 & 3.29 & \underline{5.36} \\
    DCTNet & 2.47 & 4.13 & 5.38 & 8.04 & 4.90 & 9.10 & 2.75 & 5.07 & 3.07 & 5.83 & 3.71 & 6.43 \\
    DADA   & 2.81 & 4.01 & 5.86 & 7.79 & 4.83 & 7.99 & 2.77 & 4.11 & 3.76 & 6.19 & 4.01 & 6.02 \\
    DuCos  & \underline{2.23} & \underline{3.44} & 4.60 & 6.90 & \textbf{4.61} & 7.37 & \textbf{2.23} & \underline{3.96} & \underline{2.67} & \underline{5.18} & \underline{3.27} & 5.37 \\
    \hline
    \textbf{WAVE} & \textbf{2.13} & \textbf{3.17} & \textbf{3.49} & \textbf{5.33} & \textbf{4.61} & \textbf{7.16} & \underline{2.28} & \textbf{3.52} & \textbf{2.35} & \textbf{4.86} & \textbf{2.97} & \textbf{4.81} \\
    \hline
    
	\end{tabular}
	\label{tab:hypersim}
\end{table*}

\begin{table}[!htb]
    \caption {NYU\_v2 training protocol at $8 \times$ and $16 \times$, evaluated on RGB-D-D, DIML, NYU\_v2, Middlebury, and Lu. At these scales, the benchmark is largely saturated, where the leading methods fall within roughly 0.1--0.2cm RMSE. Baselines taken from~\cite{wang2024scene, kang2025c2pd}. Best in \textbf{bold}, second-best \underline{underlined}.}
	\label{tab:nyu_v2}
	\centering
    \footnotesize
    \setlength{\tabcolsep}{2pt}
	\begin{tabular}{l|cc|cc|cc|cc|cc}
    
\hline

    \multirow{2}{*}{Method} 
      & \multicolumn{2}{c|}{\textbf{RGB-D-D}}
      & \multicolumn{2}{c|}{\textbf{DIML}}
      & \multicolumn{2}{c|}{\textbf{NYU\_v2}}
      & \multicolumn{2}{c|}{\textbf{M-bury}}
      & \multicolumn{2}{c}{\textbf{Lu}}
\\
     & $8\times$ & $16\times$
     & $8\times$ & $16\times$
     & $8\times$ & $16\times$
     & $8\times$ & $16\times$
     & $8\times$ & $16\times$ \\
    \hline
    
    DJFR   & 5.57 & 7.99 & 2.34 & 4.13 & 4.94 & 9.18 & 3.19 & 5.57 & 3.57 & 6.77 \\
    JIIF   & 1.79 & 2.87 & 1.86 & 3.22 & 2.76 & 5.27 & 1.82 & 3.31 & 1.73 & 4.16 \\
    DKN    & 1.96 & 3.42 & 1.86 & 3.22 & 3.26 & 6.51 & 2.12 & 4.24 & 2.16 & 5.11 \\
    FDSR   & 1.82 & 3.06 & 1.71 & 2.87 & 3.18 & 5.86 & 2.08 & 4.39 & 2.19 & 5.00 \\
    DCTNet & 1.74 & 3.05 & 1.71 & 3.73 & 3.16 & 5.84 & 2.05 & 4.19 & 1.85 & 4.39 \\
    DADA   & 1.83 & 2.80 & 1.71 & 2.65 & 2.74 & 4.80 & 2.03 & 4.18 & 1.87 & 4.01 \\
    SSDNet & 1.72 & 2.92 & 1.83 & 3.21 & 3.14 & 5.86 & 1.91 & 4.02 & 1.82 & 4.77 \\
    SGNet  & 1.64 & 2.55 & 1.75 & 2.71 & 2.44 & 4.77 & 1.64 & 2.95 & 1.61 & 3.55 \\
    DORNet & 1.80 & 2.97 & 1.88 & 3.21 & 2.70 & 5.60 & 1.76 & 3.48 & 1.75 & 4.41 \\
    SPFNet & 1.71 & 2.53 & - & - & \textbf{2.36} & \underline{4.55} & \underline{1.57} & \underline{2.79} & 1.56 & \underline{3.20} \\
    C2PD   & 1.63 & \underline{2.41} & 1.68 & 2.59 & \textbf{2.36} & \textbf{4.48} & \underline{1.57} & 2.80 & 1.53 &  \textbf{3.11} \\
    NAIMA  & \underline{1.61} & 2.47 & \underline{1.65} & \underline{2.52} & \underline{2.39} & 4.61 & 1.62 & \textbf{2.75} & \underline{1.44} & 3.34 \\
    % LapNet & \textbf{1.56} & 2.45 & \textbf{1.52} & \underline{2.52} & \textbf{2.33} & \underline{4.55} & - & - & - & - \\
    \hline
    \textbf{WAVE} 
    & \textbf{1.58} & \textbf{2.39} & \textbf{1.62} & \textbf{2.50} & 2.50 & 4.60 & \textbf{1.56} & 2.82 & \textbf{1.40} & \underline{3.20} \\
    \hline

    \end{tabular}
	
\end{table}

\begin{table}[!htb]
    \caption{Model complexity at $16\times$ for single-batch inference, measured on one NVIDIA RTX 4090. Trainable and frozen parameters in millions (M), FLOPs in Giga-operations (G), peak GPU memory in GB. WAVE's frozen DINOv3 backbone is not updated during training.}
    \centering
    \footnotesize
    \setlength{\tabcolsep}{3pt}
    \begin{tabular}{l|c|c|c|c}
    \hline
    Method
      & \textbf{Trainable (M)}
      & \textbf{Frozen (M)}
      & \textbf{FLOPs (G)}
      & \textbf{Memory (GB)} \\
    \hline
    DCTNet & 0.48   & --    & 5.73     & 1.36 \\
    DADA   & 31.03  & --    & 212.58   & 4.96 \\
    DuCos  & 34.38  & --    & 2927.49  & 9.02 \\
    SPFNet & 31.10  & --    & 3259.42  & 2.23 \\
    SGNet  & 85.94  & --    & 9715.16  & 1.97 \\
    C2PD   & 65.05  & --    & 414.75   & 1.27 \\
    NAIMA  & 119.93 & 21.94 & 11374.36 & 3.97 \\
    \hline
    \textbf{WAVE} & 39.69 & 21.60 & 1736.76 & 1.09 \\
    \hline
    \end{tabular}
    \label{tab:params}
\end{table}

\subsection{Qualitative Comparisons}
\label{sec:qualitative}
\begin{figure*}[!htb]
\centering
\includegraphics[width=0.95\textwidth]{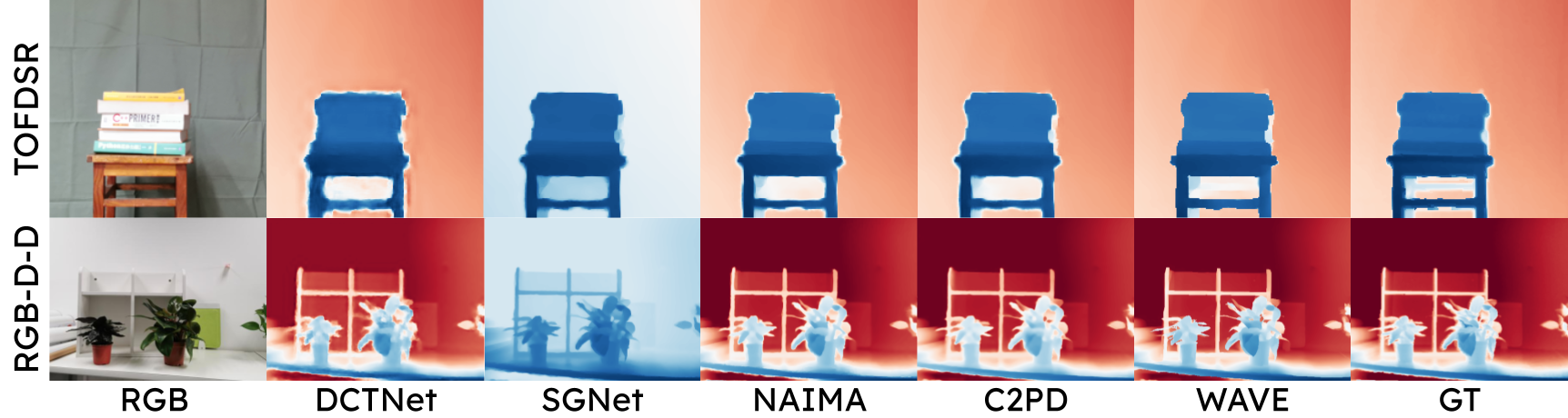}
\caption{Qualitative comparison at 8$\times$ on TOFDSR (top) and RGB-D-D (bottom). WAVE recovers thin structures and object openings with sharper, non-bleeding boundaries, closest to the ground truth (GT).}
\label{fig:qual_8x}
\end{figure*}

\begin{figure*}[!htb]
\centering
\includegraphics[width=0.95\textwidth]{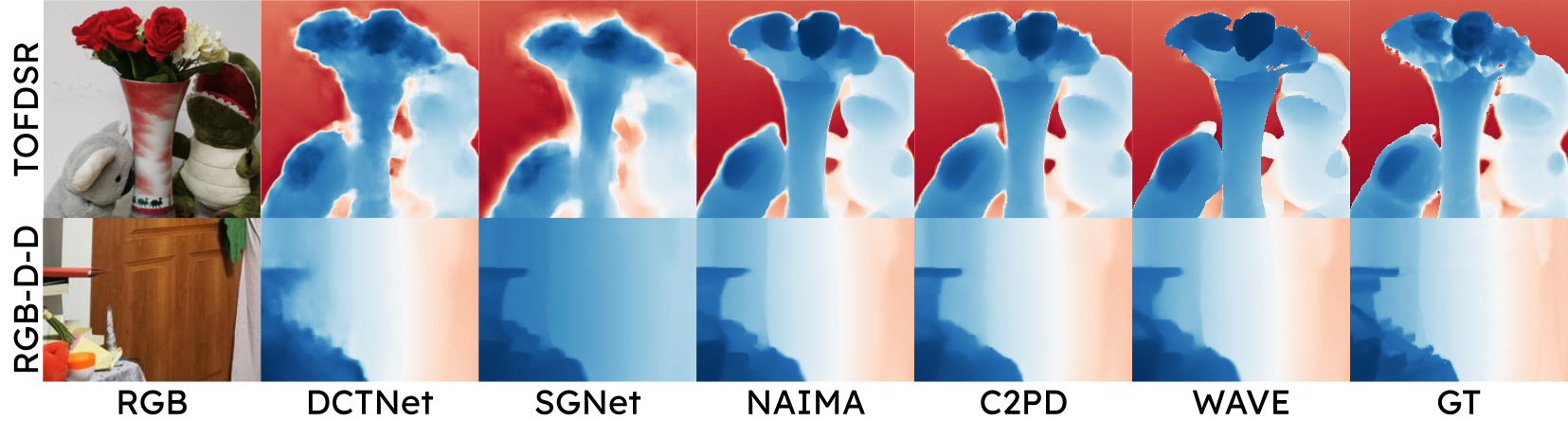}
\caption{Qualitative comparison at 16$\times$ on TOFDSR (top) and RGB-D-D (bottom). WAVE recovers sharper object boundaries and cleaner surfaces with fewer texture-copying artifacts, closest to the ground truth (GT).}
\label{fig:qual_16x}
\end{figure*}

\begin{figure}[!htb]
\centering
\includegraphics[width=0.97\columnwidth]{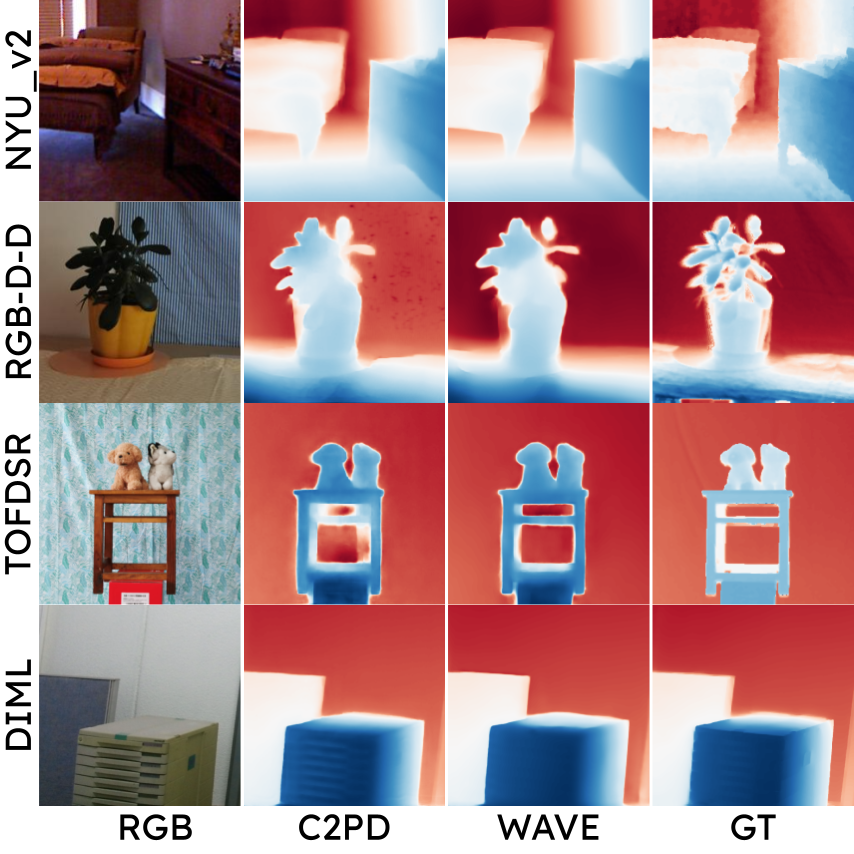}
\caption{Qualitative comparison at 32$\times$ on NYU\_v2, RGB-D-D, TOFDSR, and DIML. We compare against C2PD, the only competing method with publicly available 32$\times$ weights. WAVE preserves object openings, produces straighter boundaries, and avoids the false discontinuities and background artifacts visible in C2PD, remaining closest to GT.}
\label{fig:qual_32x}
\end{figure}

Figures~\ref{fig:qual_8x}, \ref{fig:qual_16x}, and~\ref{fig:qual_32x} show qualitative comparisons at 8$\times$, 16$\times$, and 32$\times$. All models are trained under the NYU\_v2 protocol, so TOFDSR and RGB-D-D are out-of-distribution test sets for every method shown. At 8$\times$, WAVE recovers the thin legs and crossbars of the stool on TOFDSR and preserves the gap beneath the seat, where competing methods produce blurred boundaries. At 16$\times$, WAVE produces sharper object boundaries, fewer texture-copying artifacts, and less boundary bleeding. At 32$\times$, where the low-resolution input retains the least structure and coarse-to-fine reconstruction contributes most, WAVE produces straighter object boundaries and cleaner surfaces. We compare against C2PD only at this scale, as it is the sole competing method with publicly available 32$\times$ weights.

\begin{figure}[!htb]
\centering
\includegraphics[width=0.97\columnwidth]{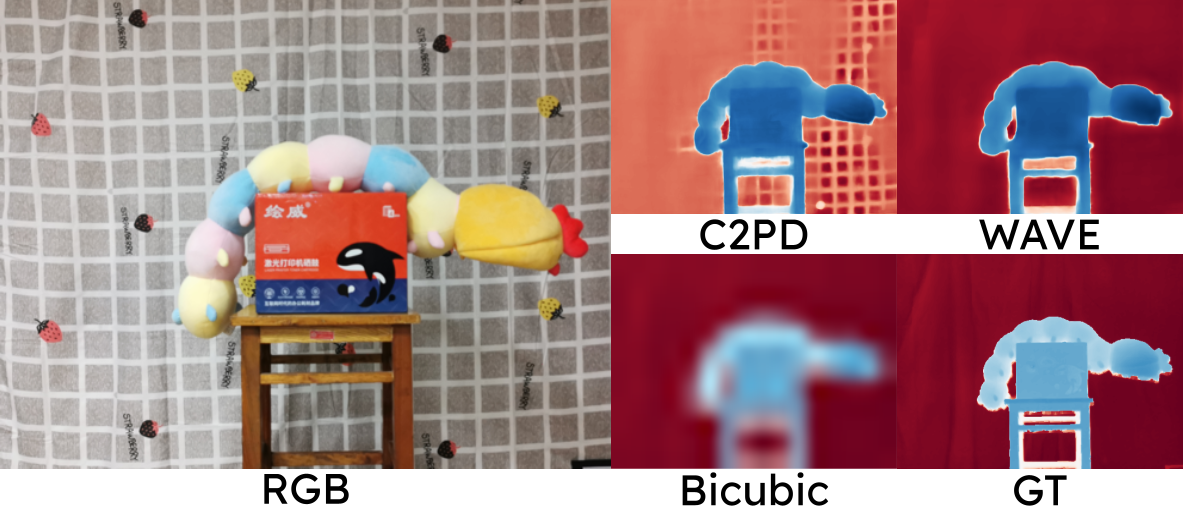}
\caption{Texture copying at 32$\times$ on TOFDSR. The grid pattern on the background cloth has no depth counterpart. C2PD transfers it into the depth map, whereas WAVE suppresses it while retaining the object boundaries.}
\label{fig:texture_copy}
\end{figure}

Figure~\ref{fig:texture_copy} shows the texture copying problem that motivates WAVE. The cloth background carries a strong grid pattern that has no counterpart in depth. C2PD copies the grid into the reconstructed depth, whereas WAVE keeps the background flat and preserves the silhouette of the toy and the stool. At 32$\times$, where most of the structure must come from RGB, the semantic gating lets WAVE separate true object boundaries from the grid, which C2PD, the current best method at this scale, fails to do. This is what WAVE is designed for. The grid is separated into the high-frequency sub-bands before any feature is learned, where semantic gating controls what is admitted, instead of entering a learned encoder that has to suppress it later.

\subsection{Robustness to RGB--Depth Misalignment}
\label{sec:shift_tests}
\begin{figure}[!htb]
\centering
\includegraphics[width=0.97\columnwidth]{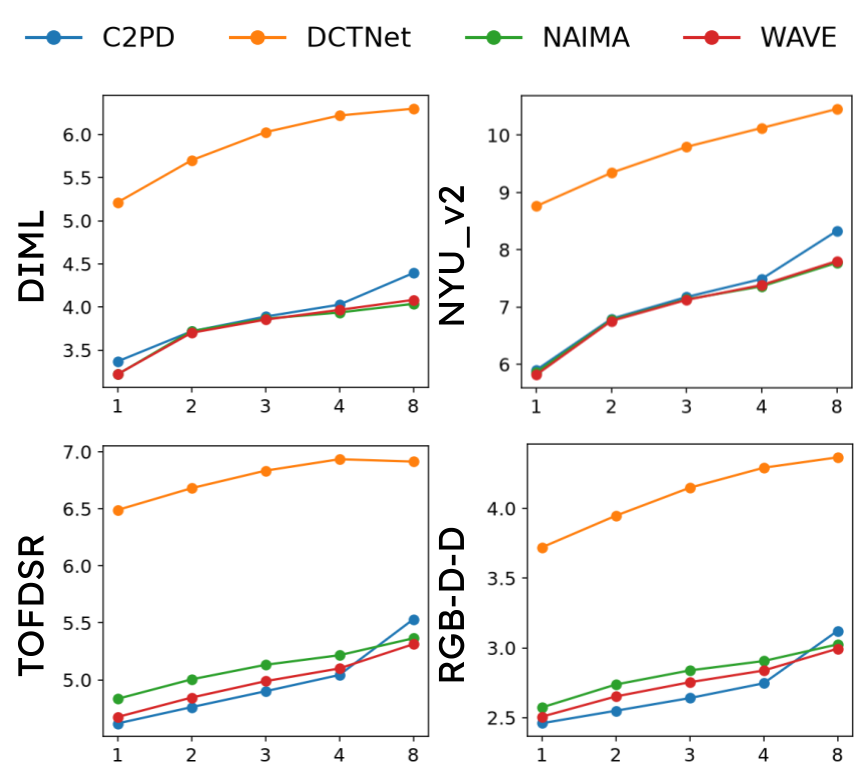}
\caption{Robustness to RGB--depth misalignment at 16$\times$. RMSE (cm, lower is better) on DIML, NYU\_v2, TOFDSR, and RGB-D-D as the RGB guide is translated by 1--8 pixels while the low-resolution depth and ground truth remain fixed.}
\label{shift}
\end{figure}

Guided methods can over-rely on pixel-accurate registration between the guide and the depth, which real RGB-D sensors rarely provide. To measure this dependence, we translate the RGB guide horizontally and vertically by 1--8 pixels on the high-resolution grid while keeping the low-resolution depth and the ground truth fixed, so any change in RMSE reflects only a method's reliance on exact alignment. We compare the 16$\times$ NYU\_v2-protocol models of DCTNet, NAIMA, C2PD, and WAVE on DIML, NYU\_v2, TOFDSR, and RGB-D-D. As Figure~\ref{shift} shows, all methods degrade as the shift grows, but WAVE remains the lowest or on par with the best competing method on all four test sets across the full range, while C2PD degrades sharply at the largest shift. 

\section{Ablation}
\label{ablations}
All ablations use WAVE trained under the NYU\_v2 protocol at $16\times$, unless stated otherwise, each isolating one component with all other settings fixed. Component ablations are evaluated on TOFDSR (560 samples), which is unseen during training and thus out-of-distribution.

\subsection{Component Ablation}
Table~\ref{tab:ablation} reports the RMSE of each variant. We start by validating our core design choice of coarse-to-fine reconstruction by reordering the ML-DWT sub-bands and layer-wise DINOv3 tokens into a fine-to-coarse schedule (WAVE-a). The increase in RMSE indicates that the gains stem not merely from the presence of the wavelet and semantic hierarchies but from how both are consumed in a structure-first order. WAVE-b replaces the IRN-style coupling with a plain channel concatenation followed by a $3 \times 3$ convolution, and the resulting drop in performance highlights the contribution of the invertible coupling. Next, we drop all wavelet sub-bands, including the LL approximation, letting the reconstruction rely solely on depth features and semantics (WAVE-c), where the drop in performance validates the role of spatial RGB guidance for both structure and detail, which semantic tokens alone cannot supply. WAVE-d removes the detail branch and the residual detail injection, leaving only low-frequency structural features and ignoring the edge and texture cues carried by the high-frequency sub-bands, which also degrades performance. WAVE-e and WAVE-f validate the role of semantics-based refinement and gating, where the former removes the semantic refinement from the structure branch, and the latter removes the gated semantic control over the high-frequency sub-bands. Removing either increases RMSE. WAVE-g removes SToRA, relying on raw DINOv3 tokens without adaptation, and WAVE-h removes the semantic boundary map derived from the final-layer DINOv3 features for object-level contour sharpening, each resulting in a performance drop. Finally, WAVE-i keeps the reversed coarse-to-fine schedule but replaces the ML-DWT pyramid with a learned CNN encoder whose four feature levels are consumed deepest first, so that the order is held fixed and only the source of the guidance changes. The drop in performance shows that reversing the order of a learned hierarchy is not enough on its own, as a CNN encoder still lets texture and color cues into its features at every level, whereas the wavelet decomposition fixes what each stage receives and keeps the high-frequency cues out of the structure path.

\begin{table}[!htb]
\caption{Ablations at $16 \times$ under the NYU\_v2 protocol. We report the RMSE in centimeters (lower is better), with the overall increase highlighting the significance of each component. Results are reported for the TOFDSR dataset, serving as an out-of-distribution test.}
\centering
\footnotesize
\setlength{\tabcolsep}{2pt}
\renewcommand{\arraystretch}{1.1}
\begin{tabular}{l|l|c}
     \hline
     Variant & \textbf{Ablated component} & \textbf{TOFDSR} \\
     \hline
     \textbf{WAVE} & - & \textbf{4.57} \\
     WAVE-a & Fine-to-coarse schedule & 4.79 \\
     WAVE-b & No cross-modal invertible coupling & 4.63 \\
     WAVE-c & No wavelet spatial bands (semantics only) & 4.73 \\
     WAVE-d & No high-frequency details injection & 4.65 \\
     WAVE-e & No semantics refinement in structure branch & 4.69 \\
     WAVE-f & No semantic gating in details branch & 4.60 \\
     WAVE-g & No SToRA & 4.62 \\
     WAVE-h & No semantic boundary refinement & 4.71 \\
     WAVE-i & \begin{tabular}[c]{@{}l@{}}Learned CNN pyramid in place of ML-DWT,\\ same reversed schedule\end{tabular} & 4.77 \\
     \hline
\end{tabular}
\label{tab:ablation}
\end{table}

\subsection{Semantic Encoder Substitution}
\label{sec:backbone_ablation}
WAVE-c shows that WAVE does not rely solely on the semantic encoder for its gains. To verify that the gains are also not tied to a specific foundation model, we replace DINOv3 with SAM-2 and ResNet-50 backbones, in each case extracting features from four layers of increasing abstraction and consuming them in the same coarse-to-fine order. Table~\ref{tab:encoders} shows that all three encoders achieve comparable performance, with DINOv3 best on RGB-D-D, Middlebury, and Lu, and SAM-2 marginally better on DIML and NYU\_v2. We adopt DINOv3 as the default since it achieves the best average performance with the fewest frozen parameters.

\begin{table}[!htb]
\caption{Semantic encoder substitution under the NYU\_v2 training protocol at $16 \times$, evaluated on RGB-D-D, DIML, NYU\_v2, Middlebury, and Lu (RMSE in cm, lower is better). Each encoder is kept frozen, and its layer-wise features are consumed in the same coarse-to-fine schedule. Best in \textbf{bold}, second-best \underline{underlined}. The last column lists frozen encoder parameters in millions (M).}
	\centering
	\footnotesize
	\setlength{\tabcolsep}{2pt}
	\begin{tabular}{l|c|c|c|c|c|c}
	\hline
	Encoder
	  & RGB-D-D
	  & DIML
	  & NYU\_v2
	  & M-bury
	  & Lu
	  & \begin{tabular}{@{}c@{}}Frozen\\Parameters\\(M)\end{tabular}
	\\
	\hline
	DINOv3
	& \textbf{2.39} & \underline{2.50} & \underline{4.60} & \textbf{2.82} & \textbf{3.20} & \textbf{21.6} \\
	SAM-2
	& \underline{2.43} & \textbf{2.47} & \textbf{4.55} & 2.87 & \underline{3.32} & 26.9 \\
	ResNet-50
	& 2.44 & 2.50 & 4.63 & \textbf{2.82} & 3.40 & \underline{23.5} \\
	\hline
	\end{tabular}
	
	\label{tab:encoders}
\end{table}

\subsection{Verifying Cross-Modal Fusion}
\label{sec:modality}
The invertible coupling is intended to keep the RGB structural signal recoverable within the fusion rather than letting the network collapse onto the depth stream. WAVE-b showed that replacing the invertible coupling with plain concatenation degrades performance, but a drop in RMSE alone does not tell whether the coupling actually keeps the RGB structural signal alive within the fusion. We test this directly to validate our hypothesis. Since the LL band is the only RGB input to this coupling, we set the LL band to zero at inference time, leaving the detail bands, semantic tokens, and depth input unchanged, and measure the change in RMSE against the unmodified model. If the coupling had collapsed onto depth, zeroing LL would leave RMSE unchanged. Table~\ref{tab:rgb_reliance} reports this test for the NYU\_v2-protocol models at $8\times$, $16\times$, and $32\times$ on the in-domain NYU\_v2 test set and the out-of-distribution DIML, RGB-D-D, and TOFDSR sets. Zeroing LL increases RMSE in every setting, showing that the fused structure depends on the RGB signal, and the increase grows with the scaling factor on all four datasets, consistent with our hypothesis that more structure must come from RGB when the low-resolution depth retains less. This empirically supports our claim that the invertible coupling fuses the two modalities without discarding either, rather than functioning as a plain mixing step that could suppress the RGB stream.

\begin{table}[!htb]
    \caption{Dependence of the fused structure on the LL band at $8\times$, $16\times$, and $32\times$ under the NYU\_v2 protocol. RMSE in centimeters (lower is better) with the LL band intact (Base) and set to zero at inference (Zero LL), and the resulting increase ($\Delta$).}
	\centering
	\footnotesize
	\setlength{\tabcolsep}{4pt}
	\begin{tabular}{c|c|c|c|c}
	\hline
	Scale
	  & Dataset
	  & Base RMSE
	  & Zero LL-band RMSE
	  & $\Delta$ RMSE
	\\
	\hline
					
    8 & DIML & 1.62 & 2.11 & 0.49 \\
    16 & DIML & 2.50 & 3.09 & 0.59 \\
    32 & DIML & 4.63 & 5.28 & 0.65 \\
    8 & NYU\_v2 & 2.50 & 3.53 & 1.03 \\
    16 & NYU\_v2 & 4.60 & 6.50 & 1.90 \\
    32 & NYU\_v2 & 7.90 & 9.93 & 2.03 \\
    8 & RGB-D-D & 1.58 & 1.66 & 0.08 \\
    16 & RGB-D-D & 2.39 & 2.66 & 0.27 \\
    32 & RGB-D-D & 3.77 & 4.17 & 0.40 \\
    8 & TOFDSR & 3.22 & 3.27 & 0.05 \\
    16 & TOFDSR & 4.57 & 4.86 & 0.29 \\
    32 & TOFDSR & 6.25 & 6.61 & 0.36 \\
	\hline
	\end{tabular}
	
	\label{tab:rgb_reliance}
\end{table}

\section{Conclusion}
We presented WAVE, a coarse-to-fine architecture for GDSR that counters the fine-to-coarse bias of conventional pipelines. By decomposing the RGB guide with a multi-level discrete wavelet transform and consuming its sub-bands together with hierarchical DINOv3 tokens in reverse order, WAVE reconstructs global structure first and fine detail last. It controls the admission of RGB cues at the input through the wavelet decomposition and processes structural, edge, and texture components through dedicated branches. We employ semantics as a learnable gating mechanism for high-frequency content and repurpose the invertible coupling for multi-modality fusion. Ablations show that the reversed coarse-to-fine schedule, the wavelet decomposition that controls which RGB cues are admitted, the semantic refinement, and gating each play a key role in achieving optimal results. Zeroing the LL band at inference increases RMSE in every setting, and more so at higher scales, confirming that the fused structure uses the RGB signal rather than collapsing onto depth. Experiments across multiple benchmarks and evaluation protocols show that WAVE performs competitively, with the largest gains at high upsampling factors where the low-resolution input retains the least structure, and with reduced sensitivity to RGB-depth misalignment. WAVE's gains are smaller in the saturated, in-domain low-scale regime, and it depends on a frozen foundation model for semantic priors. Relaxing this dependence and extending the coarse-to-fine schedule to blind or real-world degradations are promising directions.

\bibliographystyle{IEEEtran}
\bibliography{ref}

@inproceedings{yuan2023recurrent,
  title={Recurrent structure attention guidance for depth super-resolution},
  author={Yuan, Jiayi and Jiang, Haobo and Li, Xiang and Qian, Jianjun and Li, Jun and Yang, Jian},
  booktitle={Proceedings of the AAAI Conference on Artificial Intelligence},
  volume={37},
  number={3},
  pages={3331--3339},
  year={2023}
}

@inproceedings{yuan2023structure,
  title={Structure flow-guided network for real depth super-resolution},
  author={Yuan, Jiayi and Jiang, Haobo and Li, Xiang and Qian, Jianjun and Li, Jun and Yang, Jian},
  booktitle={Proceedings of the AAAI Conference on Artificial Intelligence},
  volume={37},
  number={3},
  pages={3340--3348},
  year={2023}
}

@article{zhong2023guided,
  title={Guided depth map super-resolution: A survey},
  author={Zhong, Zhiwei and Liu, Xianming and Jiang, Junjun and Zhao, Debin and Ji, Xiangyang},
  journal={ACM Computing Surveys},
  volume={55},
  number={14s},
  pages={1--36},
  year={2023},
  publisher={ACM New York, NY}
}

@inproceedings{wang2024sgnet,
  title={Sgnet: Structure guided network via gradient-frequency awareness for depth map super-resolution},
  author={Wang, Zhengxue and Yan, Zhiqiang and Yang, Jian},
  booktitle={Proceedings of the AAAI Conference on Artificial Intelligence},
  volume={38},
  number={6},
  pages={5823--5831},
  year={2024}
}

@inproceedings{wang2025dornet,
  title={Dornet: A degradation oriented and regularized network for blind depth super-resolution},
  author={Wang, Zhengxue and Yan, Zhiqiang and Pan, Jinshan and Gao, Guangwei and Zhang, Kai and Yang, Jian},
  booktitle={Proceedings of the Computer Vision and Pattern Recognition Conference},
  pages={15813--15822},
  year={2025}
}

@inproceedings{tang2021bridgenet,
  title={Bridgenet: A joint learning network of depth map super-resolution and monocular depth estimation},
  author={Tang, Qi and Cong, Runmin and Sheng, Ronghui and He, Lingzhi and Zhang, Dan and Zhao, Yao and Kwong, Sam},
  booktitle={Proceedings of the 29th acm international conference on multimedia},
  pages={2148--2157},
  year={2021}
}

@article{oquab2023dinov2,
  title={Dinov2: Learning robust visual features without supervision},
  author={Oquab, Maxime and Darcet, Timoth{\'e}e and Moutakanni, Th{\'e}o and Vo, Huy and Szafraniec, Marc and Khalidov, Vasil and Fernandez, Pierre and Haziza, Daniel and Massa, Francisco and El-Nouby, Alaaeldin and others},
  journal={arXiv preprint arXiv:2304.07193},
  year={2023}
}

@inproceedings{scharstein2003high,
  title={High-accuracy stereo depth maps using structured light},
  author={Scharstein, Daniel and Szeliski, Richard},
  booktitle={2003 IEEE Computer Society Conference on Computer Vision and Pattern Recognition, 2003. Proceedings.},
  volume={1},
  pages={I--I},
  year={2003},
  organization={IEEE}
}

@inproceedings{lu2014depth,
  title={Depth enhancement via low-rank matrix completion},
  author={Lu, Si and Ren, Xiaofeng and Liu, Feng},
  booktitle={Proceedings of the IEEE conference on computer vision and pattern recognition},
  pages={3390--3397},
  year={2014}
}

@inproceedings{silberman2012indoor,
  title={Indoor segmentation and support inference from rgbd images},
  author={Silberman, Nathan and Hoiem, Derek and Kohli, Pushmeet and Fergus, Rob},
  booktitle={European conference on computer vision},
  pages={746--760},
  year={2012},
  organization={Springer}
}

@inproceedings{roberts2021hypersim,
  title={Hypersim: A photorealistic synthetic dataset for holistic indoor scene understanding},
  author={Roberts, Mike and Ramapuram, Jason and Ranjan, Anurag and Kumar, Atulit and Bautista, Miguel Angel and Paczan, Nathan and Webb, Russ and Susskind, Joshua M},
  booktitle={Proceedings of the IEEE/CVF international conference on computer vision},
  pages={10912--10922},
  year={2021}
}

@article{cho2021diml,
  title={Diml/cvl rgb-d dataset: 2m rgb-d images of natural indoor and outdoor scenes},
  author={Cho, Jaehoon and Min, Dongbo and Kim, Youngjung and Sohn, Kwanghoon},
  journal={arXiv preprint arXiv:2110.11590},
  year={2021}
}

@inproceedings{yan2024tri,
  title={Tri-Perspective View Decomposition for Geometry-Aware Depth Completion},
  author={Yan, Zhiqiang and Lin, Yuankai and Wang, Kun and Zheng, Yupeng and Wang, Yufei and Zhang, Zhenyu and Li, Jun and Yang, Jian},
  booktitle={Proceedings of the IEEE/CVF Conference on Computer Vision and Pattern Recognition},
  pages={4874--4884},
  year={2024}
}

@article{ariav2022depth,
  title={Depth map super-resolution via cascaded transformers guidance},
  author={Ariav, Ido and Cohen, Israel},
  journal={Frontiers in Signal Processing},
  volume={2},
  pages={847890},
  year={2022},
  publisher={Frontiers Media SA}
}

@inproceedings{de2022learning,
  title={Learning graph regularisation for guided super-resolution},
  author={De Lutio, Riccardo and Becker, Alexander and D'Aronco, Stefano and Russo, Stefania and Wegner, Jan D and Schindler, Konrad},
  booktitle={Proceedings of the IEEE/CVF conference on computer vision and pattern recognition},
  pages={1979--1988},
  year={2022}
}

@inproceedings{hui2016depth,
  title={Depth map super-resolution by deep multi-scale guidance},
  author={Hui, Tak-Wai and Loy, Chen Change and Tang, Xiaoou},
  booktitle={European conference on computer vision},
  pages={353--369},
  year={2016},
  organization={Springer}
}

@article{zuo2021mig,
  title={MIG-Net: Multi-scale network alternatively guided by intensity and gradient features for depth map super-resolution},
  author={Zuo, Yifan and Wang, Hao and Fang, Yuming and Huang, Xiaoshui and Shang, Xiwu and Wu, Qiang},
  journal={IEEE Transactions on Multimedia},
  volume={24},
  pages={3506--3519},
  year={2021},
  publisher={IEEE}
}

@inproceedings{song2020channel,
  title={Channel attention based iterative residual learning for depth map super-resolution},
  author={Song, Xibin and Dai, Yuchao and Zhou, Dingfu and Liu, Liu and Li, Wei and Li, Hongdong and Yang, Ruigang},
  booktitle={Proceedings of the ieee/cvf conference on computer vision and pattern recognition},
  pages={5631--5640},
  year={2020}
}

@article{yang2022codon,
  title={CODON: On orchestrating cross-domain attentions for depth super-resolution},
  author={Yang, Yuxiang and Cao, Qi and Zhang, Jing and Tao, Dacheng},
  journal={International Journal of Computer Vision},
  volume={130},
  number={2},
  pages={267--284},
  year={2022},
  publisher={Springer}
}

@article{zhong2021high,
  title={High-resolution depth maps imaging via attention-based hierarchical multi-modal fusion},
  author={Zhong, Zhiwei and Liu, Xianming and Jiang, Junjun and Zhao, Debin and Chen, Zhiwen and Ji, Xiangyang},
  journal={IEEE Transactions on Image Processing},
  volume={31},
  pages={648--663},
  year={2021},
  publisher={IEEE}
}

@article{yan2022learning,
  title={Learning complementary correlations for depth super-resolution with incomplete data in real world},
  author={Yan, Zhiqiang and Wang, Kun and Li, Xiang and Zhang, Zhenyu and Li, Guangyu and Li, Jun and Yang, Jian},
  journal={IEEE transactions on neural networks and learning systems},
  volume={35},
  number={4},
  pages={5616--5626},
  year={2022},
  publisher={IEEE}
}

@article{nasir2026implicit,
  title={Implicit Neural Representation-Based Continuous Single Image Super Resolution: An Empirical Study},
  author={Nasir, Tayyab and Liu, Daochang and Mian, Ajmal},
  journal={arXiv preprint arXiv:2601.17723},
  year={2026}
}

@article{kim2021deformable,
  title={Deformable kernel networks for joint image filtering},
  author={Kim, Beomjun and Ponce, Jean and Ham, Bumsub},
  journal={International Journal of Computer Vision},
  volume={129},
  number={2},
  pages={579--600},
  year={2021},
  publisher={Springer}
}

@inproceedings{he2021towards,
  title={Towards fast and accurate real-world depth super-resolution: Benchmark dataset and baseline},
  author={He, Lingzhi and Zhu, Hongguang and Li, Feng and Bai, Huihui and Cong, Runmin and Zhang, Chunjie and Lin, Chunyu and Liu, Meiqin and Zhao, Yao},
  booktitle={Proceedings of the ieee/cvf conference on computer vision and pattern recognition},
  pages={9229--9238},
  year={2021}
}

@inproceedings{tang2021joint,
  title={Joint implicit image function for guided depth super-resolution},
  author={Tang, Jiaxiang and Chen, Xiaokang and Zeng, Gang},
  booktitle={Proceedings of the 29th acm international conference on multimedia},
  pages={4390--4399},
  year={2021}
}

@inproceedings{zhao2022discrete,
  title={Discrete cosine transform network for guided depth map super-resolution},
  author={Zhao, Zixiang and Zhang, Jiangshe and Xu, Shuang and Lin, Zudi and Pfister, Hanspeter},
  booktitle={Proceedings of the IEEE/CVF conference on computer vision and pattern recognition},
  pages={5697--5707},
  year={2022}
}

@inproceedings{shi2022symmetric,
  title={Symmetric uncertainty-aware feature transmission for depth super-resolution},
  author={Shi, Wuxuan and Ye, Mang and Du, Bo},
  booktitle={Proceedings of the 30th ACM International Conference on Multimedia},
  pages={3867--3876},
  year={2022}
}

@inproceedings{zhao2023spherical,
  title={Spherical space feature decomposition for guided depth map super-resolution},
  author={Zhao, Zixiang and Zhang, Jiangshe and Gu, Xiang and Tan, Chengli and Xu, Shuang and Zhang, Yulun and Timofte, Radu and Van Gool, Luc},
  booktitle={Proceedings of the IEEE/CVF International Conference on Computer Vision},
  pages={12547--12558},
  year={2023}
}

@article{zhong2023deep,
  title={Deep attentional guided image filtering},
  author={Zhong, Zhiwei and Liu, Xianming and Jiang, Junjun and Zhao, Debin and Ji, Xiangyang},
  journal={IEEE Transactions on Neural Networks and Learning Systems},
  volume={35},
  number={9},
  pages={12236--12250},
  year={2023},
  publisher={IEEE}
}

@inproceedings{metzger2023guided,
  title={Guided depth super-resolution by deep anisotropic diffusion},
  author={Metzger, Nando and Daudt, Rodrigo Caye and Schindler, Konrad},
  booktitle={Proceedings of the IEEE/CVF Conference on Computer Vision and Pattern Recognition},
  pages={18237--18246},
  year={2023}
}

@inproceedings{yan2025ducos,
  title={Ducos: Duality constrained depth super-resolution via foundation model},
  author={Yan, Zhiqiang and Wang, Zhengxue and Dong, Haoye and Li, Jun and Yang, Jian and Lee, Gim Hee},
  booktitle={Proceedings of the IEEE/CVF International Conference on Computer Vision},
  pages={8361--8371},
  year={2025}
}

@article{wang2024scene,
  title={Scene Prior Filtering for Depth Super-Resolution},
  author={Wang, Zhengxue and Yan, Zhiqiang and Yang, Ming-Hsuan and Pan, Jinshan and Gao, Guangwei and Tai, Ying and Yang, Jian},
  journal={arXiv preprint arXiv:2402.13876},
  year={2024}
}

@article{nasir2026naima,
  title={NAIMA: Semantics Aware RGB Guided Depth Super-Resolution},
  author={Nasir, Tayyab and Liu, Daochang and Mian, Ajmal},
  journal={arXiv preprint arXiv:2604.04407},
  year={2026}
}

@inproceedings{kang2025c2pd,
  title={C2pd: Continuity-constrained pixelwise deformation for guided depth super-resolution},
  author={Kang, Jiahui and Cai, Qing and Tan, Runqing and Liu, Yimei and Liu, Zhi},
  booktitle={Proceedings of the AAAI Conference on Artificial Intelligence},
  volume={39},
  number={4},
  pages={4212--4220},
  year={2025}
}

@inproceedings{zhong2026dual,
  title={Dual Graph Regularized Deep Unfolding Network for Guided Depth Map Super-resolution},
  author={Zhong, Zhiwei and Chen, Peilin and Shen, Qiangqiang and Li, Bo and Wang, Shiqi},
  booktitle={Proceedings of the IEEE/CVF Conference on Computer Vision and Pattern Recognition},
  pages={16322--16332},
  year={2026}
}

@article{wu2026degmamba,
  title={DegMamba: Mamba-enhanced depth super-resolution with degradation guidance},
  author={Wu, Qiyue and Yan, Zhiqiang and Wang, Zhengxue and Yang, Jian and Li, Jun},
  journal={Pattern Recognition},
  pages={113921},
  year={2026},
  publisher={Elsevier}
}

@article{wu2024reft,
  title={Reft: Representation finetuning for language models},
  author={Wu, Zhengxuan and Arora, Aryaman and Wang, Zheng and Geiger, Atticus and Jurafsky, Dan and Manning, Christopher D and Potts, Christopher},
  journal={Advances in Neural Information Processing Systems},
  volume={37},
  pages={63908--63962},
  year={2024}
}

@article{hu2022lora,
  title={Lora: Low-rank adaptation of large language models.},
  author={Hu, Edward J and Shen, Yelong and Wallis, Phillip and Allen-Zhu, Zeyuan and Li, Yuanzhi and Wang, Shean and Wang, Liang and Chen, Weizhu and others},
  journal={Iclr},
  volume={1},
  number={2},
  pages={3},
  year={2022}
}

@inproceedings{xiao2020invertible,
  title={Invertible image rescaling},
  author={Xiao, Mingqing and Zheng, Shuxin and Liu, Chang and Wang, Yaolong and He, Di and Ke, Guolin and Bian, Jiang and Lin, Zhouchen and Liu, Tie-Yan},
  booktitle={European conference on computer vision},
  pages={126--144},
  year={2020},
  organization={Springer}
}

@article{mallat1989theory,
  title={A theory for multiresolution signal decomposition: the wavelet representation},
  author={Mallat, Stephane G},
  journal={IEEE transactions on pattern analysis and machine intelligence},
  volume={11},
  number={7},
  pages={674--693},
  year={1989},
  publisher={Ieee}
}

@inproceedings{caron2021emerging,
  title={Emerging properties in self-supervised vision transformers},
  author={Caron, Mathilde and Touvron, Hugo and Misra, Ishan and J{\'e}gou, Herv{\'e} and Mairal, Julien and Bojanowski, Piotr and Joulin, Armand},
  booktitle={Proceedings of the IEEE/CVF international conference on computer vision},
  pages={9650--9660},
  year={2021}
}

@article{yosinski2014transferable,
  title={How transferable are features in deep neural networks?},
  author={Yosinski, Jason and Clune, Jeff and Bengio, Yoshua and Lipson, Hod},
  journal={Advances in neural information processing systems},
  volume={27},
  year={2014}
}

@inproceedings{zeiler2014visualizing,
  title={Visualizing and understanding convolutional networks},
  author={Zeiler, Matthew D and Fergus, Rob},
  booktitle={European conference on computer vision},
  pages={818--833},
  year={2014},
  organization={Springer}
}

@inproceedings{kirillov2023segment,
  title={Segment anything},
  author={Kirillov, Alexander and Mintun, Eric and Ravi, Nikhila and Mao, Hanzi and Rolland, Chloe and Gustafson, Laura and Xiao, Tete and Whitehead, Spencer and Berg, Alexander C and Lo, Wan-Yen and others},
  booktitle={Proceedings of the IEEE/CVF international conference on computer vision},
  pages={4015--4026},
  year={2023}
}

@article{liu2023one,
  title={One-2-3-45: Any single image to 3d mesh in 45 seconds without per-shape optimization},
  author={Liu, Minghua and Xu, Chao and Jin, Haian and Chen, Linghao and Varma T, Mukund and Xu, Zexiang and Su, Hao},
  journal={Advances in Neural Information Processing Systems},
  volume={36},
  pages={22226--22246},
  year={2023}
}

@inproceedings{zhang2025detect,
  title={Detect anything 3d in the wild},
  author={Zhang, Hanxue and Jiang, Haoran and Yao, Qingsong and Sun, Yanan and Zhang, Renrui and Zhao, Hao and Li, Hongyang and Zhu, Hongzi and Yang, Zetong},
  booktitle={Proceedings of the IEEE/CVF International Conference on Computer Vision},
  pages={5048--5059},
  year={2025}
}

@article{simeoni2025dinov3,
  title={Dinov3},
  author={Sim{\'e}oni, Oriane and Vo, Huy V and Seitzer, Maximilian and Baldassarre, Federico and Oquab, Maxime and Jose, Cijo and Khalidov, Vasil and Szafraniec, Marc and Yi, Seungeun and Ramamonjisoa, Micha{\"e}l and others},
  journal={arXiv preprint arXiv:2508.10104},
  year={2025}
}

@inproceedings{yermakov2026deepfake,
  title={Deepfake detection that generalizes across benchmarks},
  author={Yermakov, Andrii and Cech, Jan and Matas, Jiri and Fritz, Mario},
  booktitle={Proceedings of the IEEE/CVF Winter Conference on Applications of Computer Vision},
  pages={773--783},
  year={2026}
}

@article{lin2025depth,
  title={Depth anything 3: Recovering the visual space from any views},
  author={Lin, Haotong and Chen, Sili and Liew, Junhao and Chen, Donny Y and Li, Zhenyu and Shi, Guang and Feng, Jiashi and Kang, Bingyi},
  journal={arXiv preprint arXiv:2511.10647},
  year={2025}
}

@inproceedings{katharopoulos2020transformers,
  title={Transformers are rnns: Fast autoregressive transformers with linear attention},
  author={Katharopoulos, Angelos and Vyas, Apoorv and Pappas, Nikolaos and Fleuret, Fran{\c{c}}ois},
  booktitle={International conference on machine learning},
  pages={5156--5165},
  year={2020},
  organization={PMLR}
}

@article{elfwing2018sigmoid,
  title={Sigmoid-weighted linear units for neural network function approximation in reinforcement learning},
  author={Elfwing, Stefan and Uchibe, Eiji and Doya, Kenji},
  journal={Neural networks},
  volume={107},
  pages={3--11},
  year={2018},
  publisher={Elsevier}
}

@inproceedings{haris2018deep,
  title={Deep back-projection networks for super-resolution},
  author={Haris, Muhammad and Shakhnarovich, Gregory and Ukita, Norimichi},
  booktitle={Proceedings of the IEEE conference on computer vision and pattern recognition},
  pages={1664--1673},
  year={2018}
}

@article{li2019joint,
  title={Joint image filtering with deep convolutional networks},
  author={Li, Yijun and Huang, Jia-Bin and Ahuja, Narendra and Yang, Ming-Hsuan},
  journal={IEEE transactions on pattern analysis and machine intelligence},
  volume={41},
  number={8},
  pages={1909--1923},
  year={2019},
  publisher={IEEE}
}

@inproceedings{li2016deep,
  title={Deep joint image filtering},
  author={Li, Yijun and Huang, Jia-Bin and Ahuja, Narendra and Yang, Ming-Hsuan},
  booktitle={European conference on computer vision},
  pages={154--169},
  year={2016},
  organization={Springer}
}

@article{deng2020deep,
  title={Deep convolutional neural network for multi-modal image restoration and fusion},
  author={Deng, Xin and Dragotti, Pier Luigi},
  journal={IEEE transactions on pattern analysis and machine intelligence},
  volume={43},
  number={10},
  pages={3333--3348},
  year={2020},
  publisher={IEEE}
}

@inproceedings{chen2024intrinsic,
  title={Intrinsic phase-preserving networks for depth super resolution},
  author={Chen, Xuanhong and Wang, Hang and Chen, Jialiang and Feng, Kairui and Liu, Jinfan and Wang, Xiaohang and Zhang, Weimin and Ni, Bingbing},
  booktitle={Proceedings of the AAAI Conference on Artificial Intelligence},
  volume={38},
  number={2},
  pages={1210--1218},
  year={2024}
}

\vfill

\end{document}